\documentclass[10pt,twocolumn,letterpaper]{article}

\usepackage{cvpr}              
\definecolor{cvprblue}{rgb}{0.21,0.49,0.74}
\usepackage[pagebackref,breaklinks,colorlinks,allcolors=cvprblue]{hyperref}

\usepackage{booktabs} 
\usepackage{adjustbox}
\usepackage{makecell}

\def\paperID{18548} 
\def\confName{CVPR}
\def\confYear{2026}

\title{
VLA-R: Vision-Language Action Retrieval toward\\Open-World End-to-End Autonomous Driving
}

\author{
Hyunki Seong\thanks{Correspondence to\\ \tt\small hynkis@kaist.ac.kr; hynkis@gmail.com} \and Seongwoo Moon \and Hojin Ahn \and Jehun Kang \and David Hyunchul Shim \\
\and
School of Electrical Engineering, KAIST\\
}

\begin{document}
\maketitle
\begin{abstract}
Exploring open-world situations in an end-to-end manner is a promising yet challenging task due to the need for strong generalization capabilities. In particular, end-to-end autonomous driving in unstructured outdoor environments often encounters conditions that were unfamiliar during training. In this work, we present Vision-Language Action Retrieval (VLA-R), an open-world end-to-end autonomous driving (OW-E2EAD) framework that integrates open-world perception with a novel vision-action retrieval paradigm. We leverage a frozen vision-language model for open-world detection and segmentation to obtain multi-scale, prompt-guided, and interpretable perception features without domain-specific tuning. A Q-Former bottleneck aggregates fine-grained visual representations with language-aligned visual features, bridging perception and action domains. To learn transferable driving behaviors, we introduce a vision-action contrastive learning scheme that aligns vision-language and action embeddings for effective open-world reasoning and action retrieval. Our experiments on a real-world robotic platform demonstrate strong generalization and exploratory performance in unstructured, unseen environments, even with limited data. Demo videos are provided in the supplementary material.

\end{abstract}    
\section{Introduction}
\label{sec:intro}

End-to-end autonomous driving (E2EAD) has recently gained increasing attention as a unified learning framework that directly maps sensory observations to driving actions without the need for explicit perception, prediction, or planning modules \cite{chitta2022transfuser, codevilla2018end, hu2023planning}. By jointly optimizing the entire decision-making pipeline, E2EAD systems have shown remarkable potential to achieve efficient and scalable policy learning in complex environments. However, most existing methods remain confined to closed-world settings, where training and deployment conditions are largely consistent and the set of recognizable entities is predefined. Such limitations hinder the ability of current E2EAD models to generalize toward the open world, where novel objects, unseen road structures, or unstructured terrains frequently appear.

Open-world autonomous driving introduces two fundamental challenges: (1) perceptual openness, where the system must recognize and reason about unknown or unseen entities beyond fixed semantic categories, and (2) behavioral generalization, where the driving policy must adapt to diverse and previously unencountered conditions without explicit supervision. While recent advances in vision-language models (VLMs) \citep{cheng2024yolo,wang2025yoloe} have demonstrated strong zero-shot perception capabilities, it remains understudied how to integrate these open-world perceptual representations into an end-to-end driving framework capable of producing robust, adaptive, and interpretable actions in unstructured environments.

\begin{figure}[t]
\centering
\includegraphics[width=0.35\textwidth]{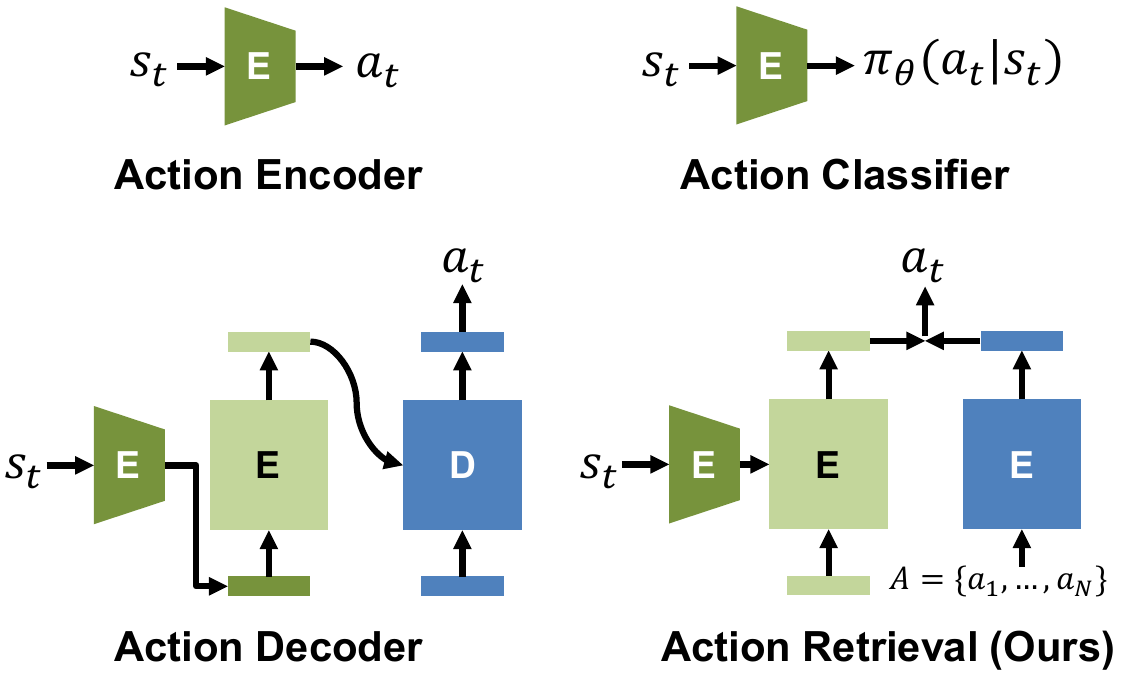}
\vspace{-5pt}
\caption{
Comparison between action generation paradigms.
}
\label{fig:compare}
\vspace{-10pt}
\end{figure}

\begin{figure*}[t]
\centering
\includegraphics[width=0.88\textwidth]{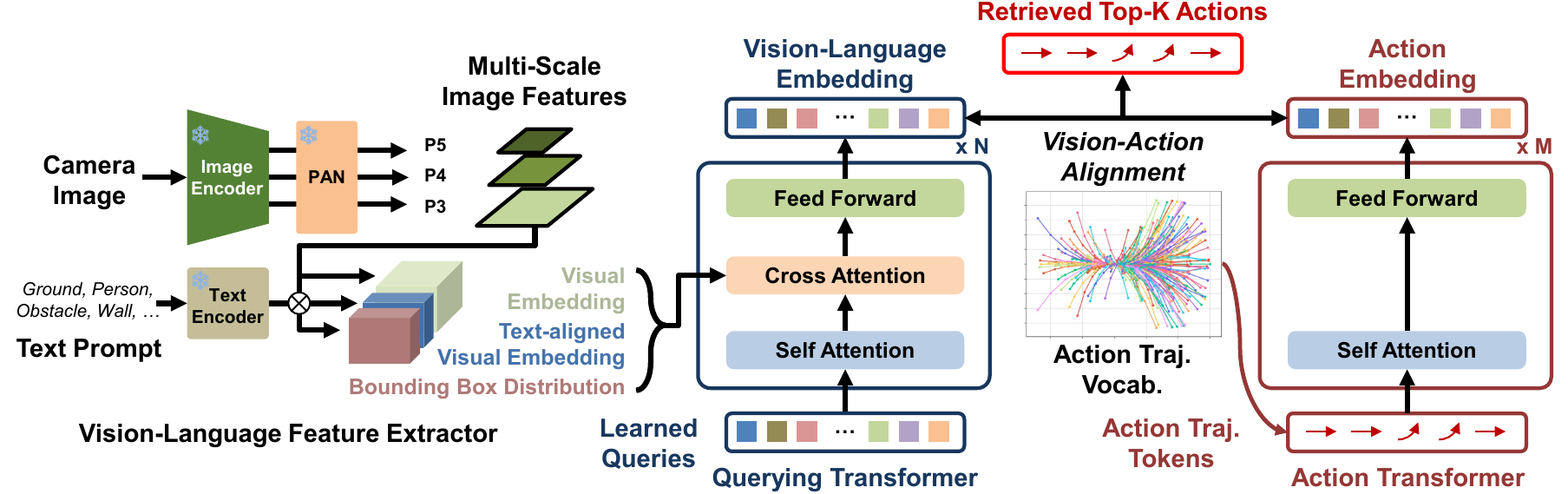}
\vspace{-5pt}
\caption{
Overview of our Vision-Language Action Retrieval (VLA-R). Our key contribution is leveraging open-world perception features together with a novel Action Retrieval mechanism, towarding generalizable autonomous driving.
}
\label{fig:overview}
\vspace{-10pt}
\end{figure*}

In this paper, we present Vision-Language Action Retrieval (VLA-R), a novel open-world end-to-end autonomous driving (OW-E2EAD) framework that bridges perception and control via a vision-action retrieval paradigm. Our key idea is to employ a frozen vision-language model to extract multi-scale, prompt-guided perception features capable of open-world object recognition and scene understanding without domain-specific fine-tuning. A Q-Former bottleneck \cite{li2023blip} aggregates fine-grained perception tokens into language-aligned visual embeddings, providing a shared latent space that links perceptual and behavioral representations. To enable transferable and interpretable decision making, we further introduce a vision-action contrastive learning scheme that aligns visual-language embeddings with corresponding driving actions, allowing the model to retrieve appropriate actions conditioned on open-world observations.

Through extensive experiments on a real-world mobile robotic platform, VLA-R demonstrates strong generalization and exploratory capability in unstructured and unseen outdoor environments, even with limited data. The results highlight the potential of vision-language integration for open-world embodied intelligence and mark a step toward scalable, interpretable, and generalizable end-to-end autonomous driving.
\section{Related Works}
\label{sec:formatting}

\noindent \textbf{End-to-End Policy Network}
Early end-to-end (E2E) driving networks primarily relied on CNN-based policy architectures to map visual observations to control commands \cite{bojarski2016end, codevilla2018end, codevilla2019exploring}. These models demonstrated effective image encoding for downstream path generation \cite{chitta2022transfuser, sridhar2024nomad}, and control prediction \cite{shahvint, wu2022trajectory} with relatively simple structures.
More recently, the introduction of large-scale datasets and powerful transformer architectures has led to Bird’s-Eye-View (BEV) transformer-based E2E \cite{chen2024vadv2, hu2023planning}, which jointly learn perception and planning through auxiliary perception heads. While these approaches achieve strong performance, their dependence on extensive annotation and high-cost data collection limits scalability.
Query-based transformer decoders have also been explored for policy learning  \citep{zhao2023learning, weng2024drive}, where learnable query embeddings represent temporal steps or motion chunks, enabling the network to autoregressively decode future actions from visual inputs. In parallel, several studies fine-tune pre-trained CNN \cite{shah2022gnm, sun2025sparsedrive, zheng2024genad} and ViT backbones \cite{hirose2025omnivla, kim2024openvla} to enhance representation quality and transferability. Despite their success, most approaches require large-scale domain-specific datasets to achieve robust generalization.
In contrast, we leverage frozen vision-language models (VLMs) \citep{liu2024grounding,zhang2024vision,wang2025yoloe} for open-world perception, enabling a generalizable perception module within our end-to-end architecture without domain-specific fine-tuning. Furthermore, instead of directly predicting action values, we propose an action retrieval paradigm that retrieves tokenized action trajectories conditioned on vision–language representations, yielding more reliable and robust performance in unseen and unstructured scenarios, even with limited training data.

\noindent \textbf{Foundation Models for Open-World Perception.}
End-to-end perception has served as a backbone architecture for learning-based E2EAD. Early approaches relied on CNN-based models with residual \cite{he2016deep} or skip-connection designs \cite{ronneberger2015u} to extract visual features. The introduction of Transformers \cite{vaswani2017attention} further advanced this direction by tokenizing images for attention-based representation learning \cite{dosovitskiy2020image}, interpretability \cite{seongself}, and enhancing spatial understanding through BEV representations \cite{li2024bevformer, liu2023bevfusion}.
Recent progress in self-supervised learning and large-scale distillation has led to visual foundation models \cite{zhang2022dino, oquab2023dinov2}, which offer highly generalizable visual features. Beyond vision-only representations, vision–language aligned perception has emerged, enabling open-vocabulary understanding via text-conditioned embeddings. CLIP \cite{radford2021learning} and subsequent open-world detectors like Grounding-DINO \cite{liu2024grounding} and YOLO-World \cite{cheng2024yolo} leverage semantic similarity to text prompts to recognize unseen object categories, moving beyond closed-set label prediction.
YOLOE \cite{wang2025yoloe} extends these ideas to unified open-world object detection and segmentation, achieving generalizable perception with real-time performance through its Re-parameterizable Region–Text Alignment (RepRTA) module.
Motivated by these advances, we employ real-time vision–language perception models capable of open-world perception as the foundation for an under-explored open-world end-to-end autonomous driving framework, enabling generalizable perception without domain-specific fine-tuning.

\section{Methodologies}

\subsection{Vision-Language Action Retrieval}
End-to-end policy networks compute driving actions directly from sensory observations under various paradigms (Fig.~\ref{fig:overview}). 
Classic continuous-control approaches model the policy as a deterministic mapping (Eq.~\ref{eq:enc}), encoding input states $s_t$ into latent features and predicting corresponding action signals $a_t$. 
Discrete-action formulations instead model a categorical distribution (Eq.~\ref{eq:cls}), improving stability but limiting expressiveness in fine-grained motion. 
Recent transformer-based frameworks adopt encoder-decoder structures (Eq.~\ref{eq:dec}) that decode actions from learnable queries. While these paradigms have advanced end-to-end control, they remain inherently \textit{closed-world}, constrained by fixed training distributions and lacking semantic awareness of unseen visual concepts or driving contexts.
\begin{subequations}
\label{eq:action_paradigms}
\begin{align}
    \textit{Action Encoder:} \quad & a_t \leftarrow f_{\text{enc}}(s_t) \label{eq:enc} \\
    \textit{Action Classifier:} \quad & a_t \sim f_{\text{cls}}(a_t|s_t) \label{eq:cls} \\
    \textit{Action Decoder:} \quad & a_t \leftarrow f_{\text{dec}}(Q, f_{\text{enc}}(s_t)) \label{eq:dec}
\end{align}
\end{subequations}

We propose \textbf{Vision-Language Action Retrieval (VLA-R)}, a novel paradigm that retrieves the most behaviorally aligned action from an \textit{open-world vision-language-action embedding space}:
\begin{equation}
    a_t \leftarrow \arg\max_A \operatorname{sim}\big(f_{query}(Q, s_t), f_{action}(A)\big),
\end{equation}
where $f_{query}$ extracts query-aligned visual embeddings from input states using a frozen vision-language model and a Q-Former bottleneck, and $f_{action}$ encodes candidate actions $A$ into the same latent space. 
Training with a vision-action contrastive loss aligns perceptual and behavioral representations, allowing the policy to retrieve semantically consistent actions without explicit decoding or classification. 
This retrieval-based paradigm bridges perception and control through open-world semantic alignment, yielding interpretable, transferable, and zero-shot generalizable action reasoning in unstructured environments.

\subsection{Open-World Querying Transformer}
\label{sec:owqformer}
\begin{figure}[t]
\centering
\includegraphics[width=0.45\textwidth]{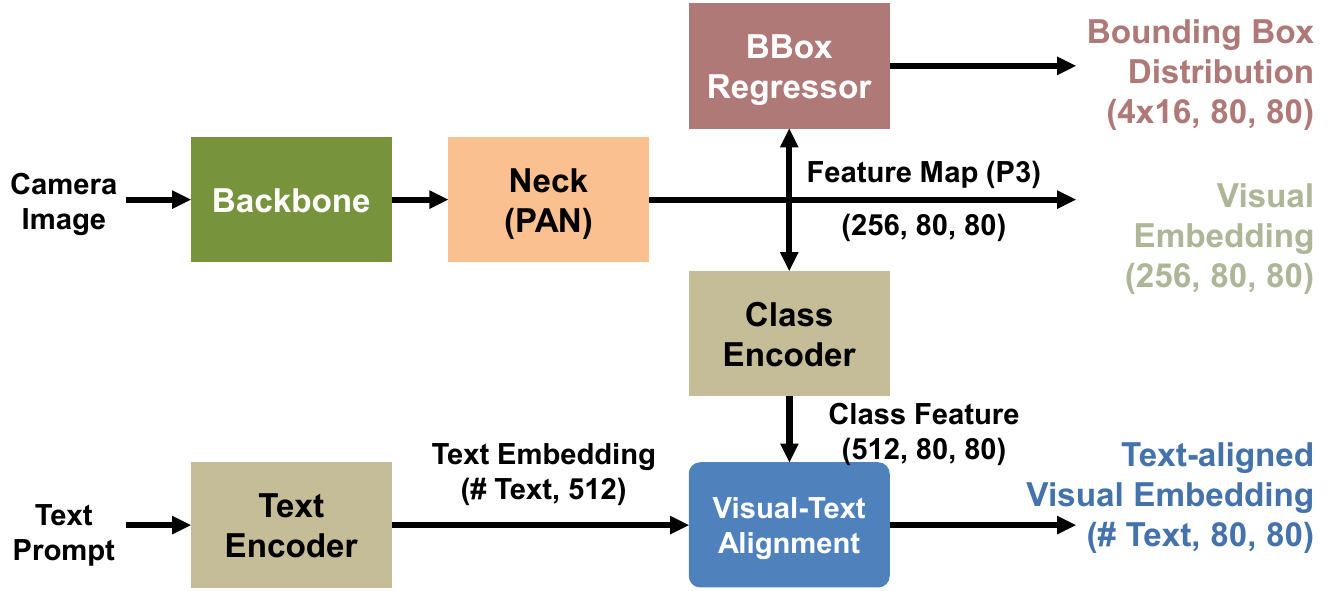}
\vspace{-5pt}
\caption{
We utilize a frozen open-world perception network as the backbone, enabling OW-E2EAD in real-world environments. In addition to visual embeddings, we use text-aligned visual features and bounding-box distributions as inputs.
}
\label{fig:yoloe}
\vspace{-15pt}
\end{figure}

%

As illustrated in Fig.~\ref{fig:overview}, the proposed \textit{Open-World Querying Transformer (OW-QFormer)} integrates multi-source embeddings from YOLOE~\cite{wang2025yoloe} into a unified, query-driven latent representation. The module bridges visual, geometric, and textual cues to form interpretable open-world scene embeddings for downstream policy reasoning.

\paragraph{Multi-Source Perception Inputs.}
YOLOE’s backbone \cite{redmon2016you} and Path Aggregation Network (PAN) \cite{liu2018path}, yield a dense feature map $\mathbf{F}_{\text{vis}} \in \mathbb{R}^{256 \times 80 \times 80}$, encoding mid-level spatial semantics such as object structure and scene layout.  
The class encoder, aligned through a frozen text encoder, produces \textit{text-aligned visual embeddings} $\mathbf{F}_{\text{txt}} \in \mathbb{R}^{N_t \times 80 \times 80},$ where $N_t$ denotes the number of text prompts. These normalized embeddings are comparable to text embeddings in the shared vision–language space, enabling open-vocabulary perception without predefined class labels.  
Meanwhile, the bounding-box regressor outputs a \textit{geometric-spatial prior} $\mathbf{F}_{\text{box}} \in \mathbb{R}^{64 \times 80 \times 80},$ where $64=4\times\texttt{reg\_max}$ represents the quantized coordinate distributions used in Distribution Focal Loss (DFL)~\cite{li2020generalized}. This prior provides object-level localization cues that complement semantic embeddings.

\paragraph{Query-Based Aggregation.}
The OW-QFormer employs $N_q$ learnable queries $\mathbf{Q} = \{\mathbf{q}_1, \ldots, \mathbf{q}_{N_q}\},$ to aggregate $\mathbf{F}_{\text{vis}}, \mathbf{F}_{\text{txt}},$ and $\mathbf{F}_{\text{box}}$ via multi-head cross-attention:
\begin{equation}
\mathbf{h}^{v}_{i,q} = \text{CrossAttn}\big(\text{SelfAttn}(\mathbf{q}_i, \textbf{Q}),\,[\mathbf{F}_{\text{vis}};\mathbf{F}_{\text{txt}};\mathbf{F}_{\text{box}}]\big)
\end{equation}
where the feature vector $\mathbf{h}^{v}_{i,q}$ is fed into an FFN to derive a fused \textit{vision-language embedding} $\mathbf{z}^{v}_{i,q}$ for each query.
Each query adaptively attends to semantically and spatially relevant regions, summarizing open-world perception into a small set of interpretable latent tokens. These compact embeddings serve as structured, language-grounded inputs for subsequent action retrieval and policy reasoning modules.

\noindent \textbf{Language Prompt.} We define a text vocabulary of \textit{\{``ground", ``grass", ``leaf", ``road", ``tree", ``person", ``stump", ``curb", ``bush"\}} covering common outdoor unstructured elements, with additional non-outdoor terms (``bike" , ``wall", ``floor floor") to approximate a prototype-scale open vocabulary. We also include the high-level class \textit{``obstacle"} to evaluate whether our OW-E2EAD framework can handle unstructured, hard-to-label objects through language-aligned features.

\subsection{Action Transformer}
\begin{figure}[t]
\centering
\includegraphics[width=0.5\textwidth]{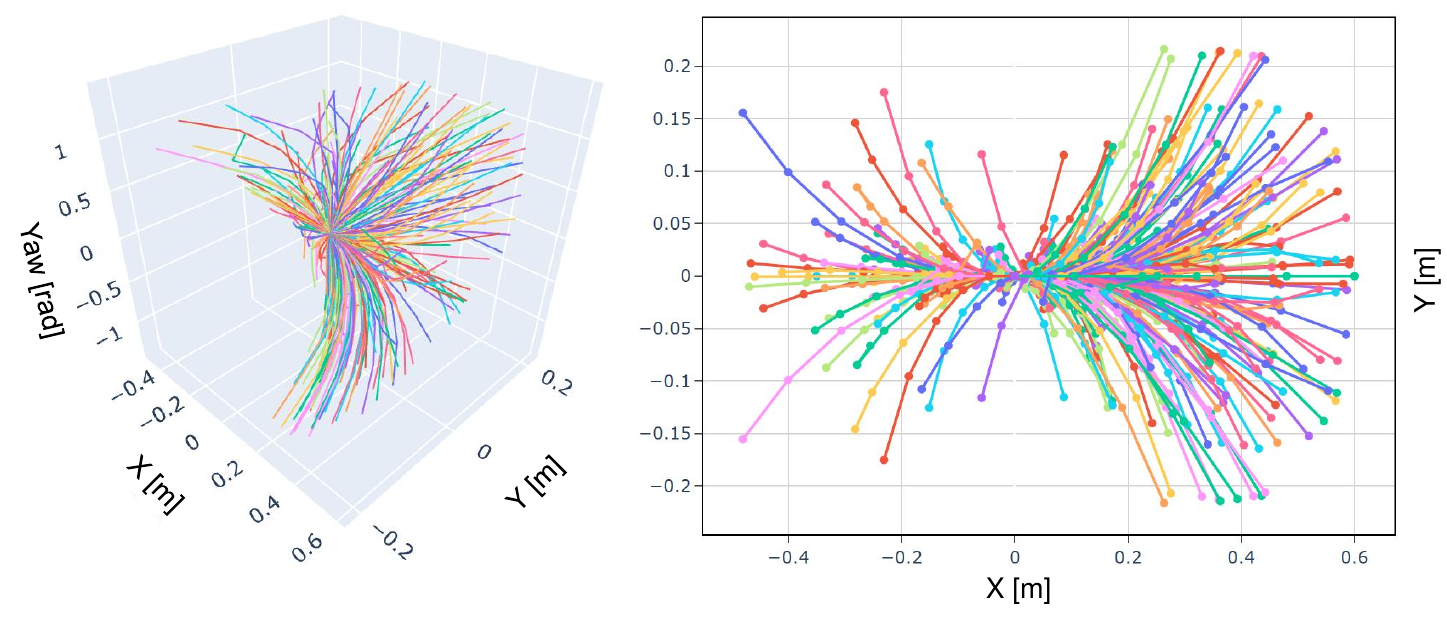}
\vspace{-5pt}
\caption{
Our three-dimensional action token vocabulary.
}
\label{fig:vocab}
\vspace{-10pt}
\end{figure}
%
%
\label{sec:action_transformer}
To reason over diverse driving behaviors, we introduce the \textit{Action Transformer}, a transformer encoder that models tokenized action trajectories within a unified latent space. Unlike prior methods that rely on fixed codebooks \citep{van2017neural} or learnable query tokens \citep{li2023blip}, our design directly encodes action trajectories, enabling flexible vocabulary construction and generalization beyond pre-defined motion primitives.
\paragraph{Architecture.}
As illustrated in Fig.~\ref{fig:overview}, the Action Transformer follows the encoder structure used for text embeddings in BLIP-2~\cite{li2023blip}, consisting of self-attention and feed-forward layers. Given a set of action trajectory tokens $\mathbf{A} = \{\mathbf{a}_1, \ldots, \mathbf{a}_M\}$, the transformer encodes them as
\begin{equation}
\mathbf{h}^{a}_i = \text{SelfAttn}(\mathbf{a}_i, \mathbf{A}), \quad
\mathbf{z}^{a}_i = \text{FFN}(\mathbf{h}\textcolor{blue}{^a}_i),
\label{eq:action_transformer}
\end{equation}
where $\mathbf{z}^{a}_i$ denotes the contextualized embedding of the $i$-th trajectory token. The output action embeddings form a structured \textit{action embedding space} used for contrastive alignment and retrieval with the vision–language embeddings from Sec.~\ref{sec:owqformer}.
\paragraph{Action Trajectory Tokenization.}
We construct an \textit{action token vocabulary} for differential-wheeled robots. Each action is represented as a six-step motion trajectory in 3D space with coordinates $(x, y, \psi)$ with corresponding control action values $(v_x, \dot{\psi})$, where $\psi$ denotes yaw angle.  
We generate six-step state rollouts from five-step control inputs using the robot’s kinematic model and discretize the resulting trajectories in a 3D grid space. For each grid cell, we compute the mean of nearby trajectories to reduce noise and obtain stable representative motions \citep{zhang2025trajtok}:
\begin{equation}
\mathcal{T}_k = \text{mean}\big\{\tau_j \mid \tau_j \in \text{cell}_k\big\},
\end{equation}
where $\mathcal{T}_k$ denotes the $k$-th trajectory token.  
Velocity commands $(v_x, \dot{\psi})$ associated with each rollout are also averaged, enabling direct retrieval of executable control signals from tokenized trajectories.
\paragraph{Action Token Encoding.}
During training, sampled trajectories are encoded by assigning each trajectory to its nearest tokenized counterpart using the $\ell_2$ distance:
\begin{equation}
\hat{k} = \arg\min_k \|\tau - \mathcal{T}_k\|_2,
\label{eq:l2token}
\end{equation}
allowing seamless tokenization and de-tokenization between continuous and discrete action spaces. Note that we compute the $\ell_2$ distance in the $(x, y, \psi)$ space to preserve fine-grained action trajectories, including rotational maneuvers. This formulation supports fine-grained and constrained motion vocabularies without re-training, enabling transferable behavior representations: dense vocabularies for precise control and constrained vocabularies for nonholonomic control schemes such as Ackermann steering.
%

\subsection{Vision-Action Representation Learning}
\label{sec:vision_action_learning}
We introduce a vision-action representation learning framework that aligns language-guided visual embeddings with action embeddings in a shared latent space.  
Inspired by \citep{radford2021learning, li2023blip}, we formulate a contrastive objective that jointly trains the \textit{Open-World Querying Transformer} (Sec.~\ref{sec:owqformer}) and the \textit{Action Transformer} (Sec.~\ref{sec:action_transformer}) to maximize correspondence between perception and motion.
%

Given a batch of $N$ paired samples $\{(\mathbf{z}^{v}_{i}, \mathbf{z}^{a}_i)\}_{i=1}^N$,  
where $\mathbf{z}^{v}_{i}=\{\mathbf{z}^{v}_{i,1}, \ldots, \mathbf{z}^{v}_{i,N_q}\}$ is the vision-language embedding from the OW-QFormer and $\mathbf{z}^{a}_i$ is the corresponding action embedding from the Action Transformer. We compute cosine similarities with max pooling across queries and apply a learnable temperature parameter~$\tau$ to all pairwise comparisons in the batch:
\begin{equation}
    s_{ij}
    =
    \tau 
    \max_{q \in \{1,\dots,N_{q}\}}
    \left(
    \mathbf{z}_{i,q}^{v\top}\mathbf{z}^{a}_{j}
    \right),
    \label{eq:similarity}
\end{equation}
Following \citep{li2023blip}, we apply the InfoNCE objective to maximize the similarity between matching pairs while minimizing it for mismatched ones:
\begin{flalign}
    \mathcal{L}_{\text{NCE}} \!=\! -\frac{1}{2N}\!\sum_{i=1}^{N} \! \Big[ \!
    \log \! \frac{\exp(s_{ii})}{\sum_{j=1}^{N}\exp(s_{ij})}
    \! + \! \log \! \frac{\exp(s_{ii})}{\sum_{j=1}^{N}\exp(s_{ji})}
    \! \Big].
    \label{eq:infonce}
\end{flalign}  
This symmetric formulation enforces bidirectional consistency: visual embeddings need to retrieve correct action embeddings, and vice versa.  
The resulting latent space preserves both semantic and geometric correlations between the perceived scene and its feasible maneuvers.
%

At inference time, a language-guided perception feature is first processed through the OW-QFormer to obtain $\mathbf{z}^{v}_{test}$, while all tokenized trajectories are encoded via the Action Transformer to produce $\{\mathbf{z}^a_1, \ldots, \mathbf{z}^a_K\}$.  
We then compute pairwise cosine similarities and perform top-\textit{k} retrieval:
\begin{equation}
\{\hat{k}_1, \ldots, \hat{k}_k\} = \text{top-}\textit{k}\big(\text{sim}(\mathbf{z}^{v}_{test}, \mathbf{a}_k)\big).
\label{eq:retrieval}
\end{equation}
In this work, we simply select the final control action as the top-1 retrieved token, though the framework can be extended with test-time scaling strategies~\citep{stiennon2020learning} to improve diversity and robustness.

By aligning vision and action in a shared contrastive space, the proposed method enables interpretable reasoning over both visual semantics and feasible motion outcomes.  
This unified embedding formulation supports open-world generalization: features from novel objects or scenarios, expressed through language-aligned representations, can be directly mapped to physically valid motion primitives without retraining.
\section{Experiments}
\begin{figure}[t]
\centering
\includegraphics[width=0.5\textwidth]{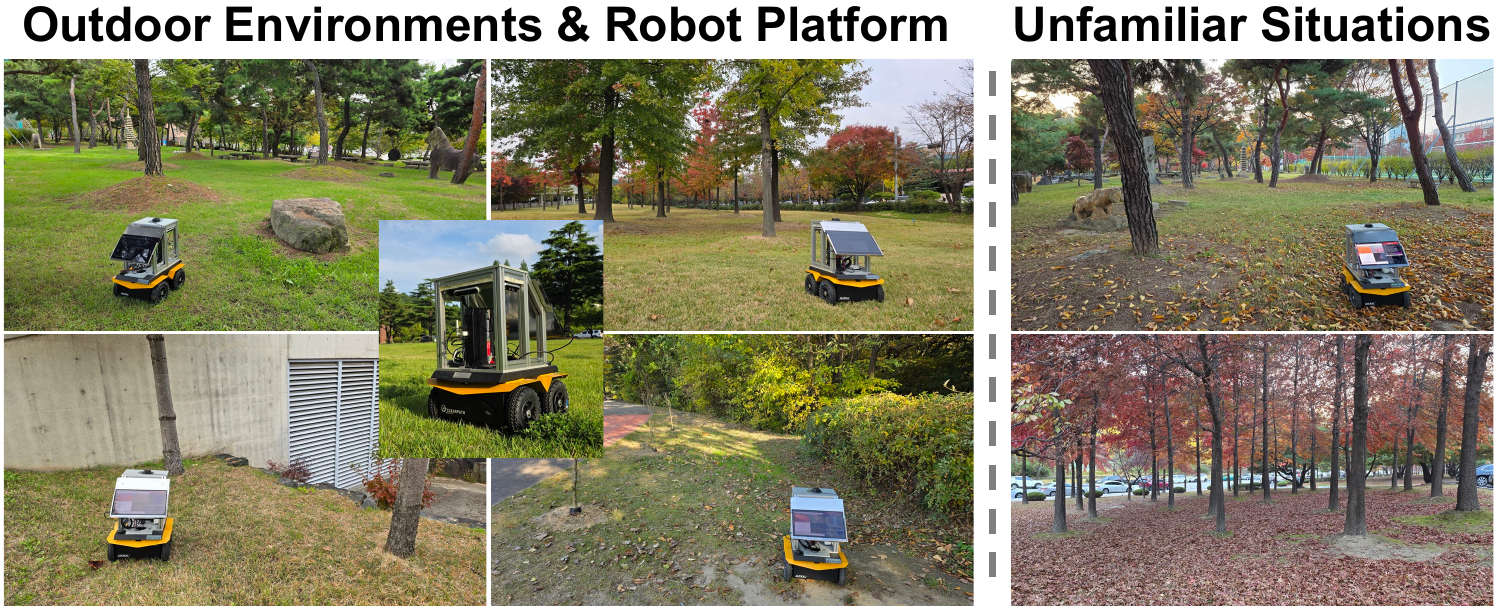}
\vspace{-10pt}
\caption{
Scenarios and robot platform used in our experiments.
}
\label{fig:experiment}
\vspace{-10pt}
\end{figure}

\subsection{Experimental Setup}
\noindent \textbf{Scenarios.}
We evaluated our approach in outdoor off-road collision-free exploration scenarios to demonstrate open-world generalization of end-to-end autonomous driving (Fig. \ref{fig:experiment}). Experiments were conducted in \textit{Rough-Terrain} scenes containing trees, stumps, rocks, and grass, highly unstructured environments that are difficult to annotate. We also evaluated two challenging cases, \textit{Cliff} and \textit{Dead-End}, which require complex turning or back-and-forth maneuvers to escape confined or hazardous areas. Each scenario included fewer than two demonstration rollouts, representing rare real-world cases. For unseen environments, we additionally tested in \textit{Dense Trees}, a setting for which no training data were collected. Data were collected and deployed across different periods of the year, introducing natural domain shifts due to seasonal and terrain variations.

\noindent \textbf{Robot Platform.}
We performed closed-loop real-world experiments on a Clearpath Jackal differential-drive robot equipped with an Intel NUC 11 Enthusiast for onboard GPU inference. The system relies solely on a front RGB camera (Intel RealSense D455) for perception, while GPS-based EKF localization provides coarse position for logging only.

\noindent \textbf{Dataset and Computational Resources.}
Training data were collected via manual control at 5 Hz, yielding 36,582 image–action pairs (near 2 h of driving). Five-step action sequences for each trajectory were tokenized using Eq.~\ref{eq:l2token}. We trained our model for 20 epochs on a single RTX 5000 Ada (32 GB) GPU, requiring less than 4 hours. Despite the limited data and compute, our framework achieved robust performance across unfamilar rough-terrain scenarios.

\subsection{Baseline Approaches}
\noindent We compare our method against representative paradigms for action generation and different visual backbones to assess both policy structure and perception generalization.

\noindent \textbf{Action Paradigms.}
We categorize policy networks by how they generate actions from sensory inputs, all sharing the same visual backbone and Q-Former encoder.
(1) \textit{Action Encoder} directly outputs continuous control signals from encoded states, similar to \citep{codevilla2018end,shah2022gnm}.
(2) \textit{Action Decoder} also predicts continuous actions but leverages a query-based decoder \cite{zhao2023learning,weng2024drive} to capture contextual dependencies and produce action \textit{chunks}.
(3) \textit{Action Classifier} discretizes the action space at the token level and selects action tokens (trajectories) by predicting class probabilities over predefined tokens.
(4) \textit{Action Retrieval} (ours) retrieves the most semantically aligned action token from a learned vocabulary, enabling interpretable and transferable reasoning in open-world conditions.

\noindent \textbf{Perception Backbones.}
We evaluate three feature extractors to study the impact of open-world perception. \textit{ResNet-101} \cite{he2016deep} serves as a strong CNN baseline, while \textit{DINO-v2} \cite{oquab2023dinov2} provides self-supervised, foundation-level visual representations with enhanced generalization. For vision-language modeling, \textit{YOLOE} \cite{wang2025yoloe} extracts prompt-guided, language-aligned embeddings. All backbones remain frozen during training to assess the strength of text-aligned, open-world perception features for OW-E2EAD.

\begin{figure}[t]
\centering
\includegraphics[width=0.48\textwidth]{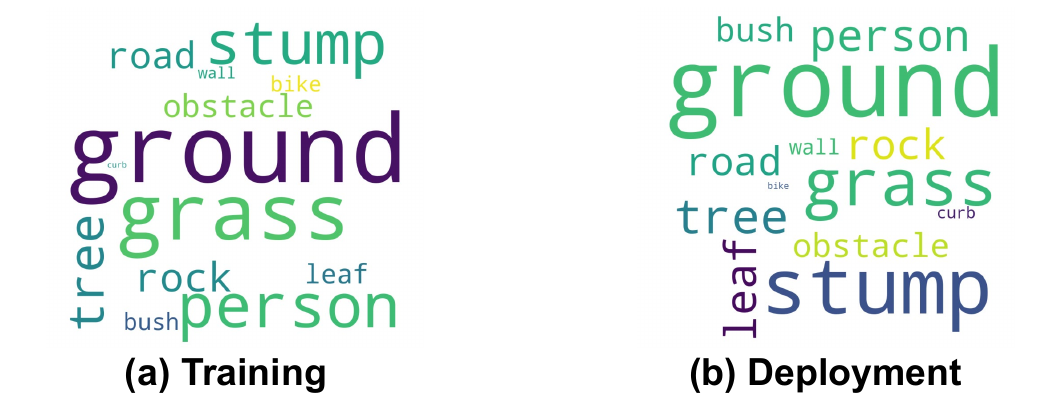}
\vspace{-5pt}
\caption{
Word clouds of training and deployment data.
}
\label{fig:result_word_cloud}
\vspace{-5pt}
\end{figure}

\begin{figure*}[t]
\centering
\includegraphics[width=0.98\textwidth]{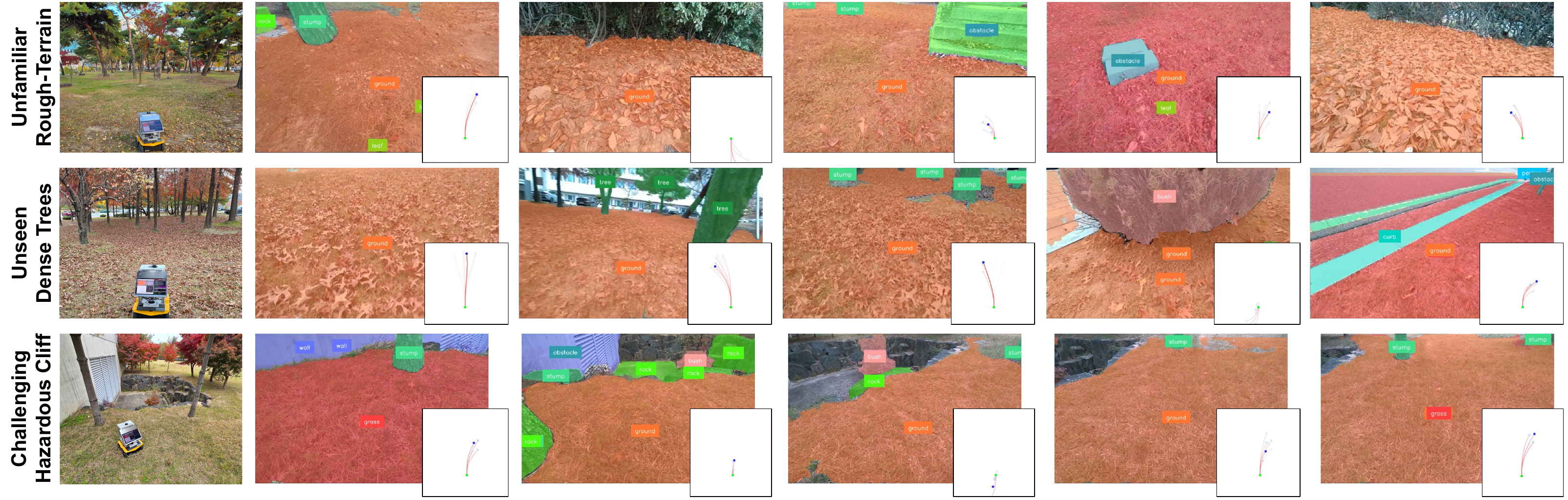}
\vspace{-5pt}
\caption{
Open-world perception and end-to-end action retrieval in diverse outdoor scenarios. Retrieved action trajectories are visualized for each perception result.
}
\label{fig:result_vla-r}
\vspace{-5pt}
\end{figure*}

\subsection{Closed-Loop Evaluation.}
We evaluate VLA-R in closed-loop real-world deployment and compare it against representative action-generation paradigms and visual backbones (Tables \ref{tab:compare_paradigm}-\ref{tab:compare_challenging}). The number of encountered collision events (\# events) reflects how often the robot encountered obstacles during exploration: more events indicate better exploration behavior. The \textit{Success} metric measures the overall collision-avoidance success rate during each experiment.
\begin{table}[t!]
  \centering
  \begin{adjustbox}{width=0.45\textwidth}
  \renewcommand{\arraystretch}{1.0}
  \begin{tabular}{@{}ccccc@{}}
    \toprule
    &\multicolumn{2}{c}{\textit{Rough-Terrain}} &\multicolumn{2}{c}{\textit{Dense-Trees}}  \\
    \cmidrule(lr){2-3} \cmidrule(lr){4-5}
    Method & \# Events & Success & \# Events & Success \\
    \midrule
    Action Encoder  & 24 & 0.79 & 13 & 0.62 \\
    Action Decoder & 24 & 0.79 & 17 & 0.71 \\
    Action Classifier & 30 & 0.83 & 40 & 0.88 \\
    Action Retrieval & 117 & 0.96 & 70 & 0.93 \\
    \bottomrule
  \end{tabular}
  \end{adjustbox}
  \caption{Comparison of Action Generation Paradigms.}
  \label{tab:compare_paradigm}
\end{table}

\noindent \textbf{Action Generation Paradigms.}
As shown in Table \ref{tab:compare_paradigm}, VLA-R achieves a large improvement over encoder-, decoder-, and classifier-based baselines. In \textit{Rough-Terrain}, VLA-R attains a success rate of 0.96 with 117 events, reflecting more extensive exploration with better collision avoidance than baselines, which show success rates between 0.79-0.83 and significantly lower event counts (13–30). In \textit{Dense Trees} (unseen environment), VLA-R maintains 0.93 success, whereas all baselines degrade, with the best achieving only 0.88. These results demonstrate that retrieval-based action selection is substantially more robust under open-world appearance shifts than direct regression or classification.

\begin{table}[t!]
  \centering
  \begin{adjustbox}{width=0.45\textwidth}
  \renewcommand{\arraystretch}{1.0}
  \begin{tabular}{@{}ccccc@{}}
    \toprule
    &\multicolumn{2}{c}{\textit{Rough-Terrain}} &\multicolumn{2}{c}{\textit{Dense-Trees}}  \\
    \cmidrule(lr){2-3} \cmidrule(lr){4-5}
    Method & \# Events & Success & \# Events & Success \\
    \midrule
    CNN (ResNet-101)  & 37 & 0.86 & 27 & 0.81 \\
    Foundation (DINO-v2) & 20 & 0.75 & 22 & 0.77 \\
    VLA-R (YOLOE) & 117 & 0.96 & 70 & 0.93 \\
    \bottomrule
  \end{tabular}
  \end{adjustbox}
  \caption{Comparison of Visual Feature Extractors.}
  \label{tab:compare_feature}
  \vspace{-5pt}
\end{table}

\noindent \textbf{Visual Perception Backbones.}
Table \ref{tab:compare_feature} compares perception modules under identical policy structures. VLA-R with the YOLOE open-world backbone consistently outperforms ResNet-101 and DINO-v2, achieving 0.96/0.93 success in Rough-Terrain/Dense Trees, compared to 0.86/0.81 (ResNet-101) and 0.75/0.77 (DINO-v2). The higher event count for our YOLOE-based method indicates that the robot explores farther while still avoiding collisions with high reliability. These results highlight the importance of text-aligned, open-world perception for stable scene understanding in unstructured environments.

\begin{table}[t!]
  \centering
  \begin{adjustbox}{width=0.3\textwidth}
  \renewcommand{\arraystretch}{1.0}
  \begin{tabular}{@{}cccc@{}}
    \toprule
    &\multicolumn{2}{c}{\textit{Challenging Scenarios}} \\
    \cmidrule(lr){2-3}
    Method & \# Events & Success \\
    \midrule
    Action Decoder & 2 & 0.1 \\
    ResNet-101 & 2 & 0.1 \\
    DINO-v2 & 11 & 0.55 \\
    VLA-R & 17 & 0.85 \\
    \bottomrule
  \end{tabular}
  \end{adjustbox}
  \caption{Results in Challenging Unstructured Scenarios.}
  \label{tab:compare_challenging}
  \vspace{-5pt}
\end{table}

\noindent \textbf{Challenging Unstructured Scenarios.}
Table \ref{tab:compare_challenging} evaluates performance in hazardous \textit{Cliff} and \textit{Dead-End} situations. VLA-R achieves 0.85 success, substantially outperforming the best baseline (DINO-v2, 0.55). Encoder-, decoder-, and ResNet-based policies frequently fail (0.1 success) due to limited generalization and inability to handle narrow passages or highly irregular geometry. In contrast, VLA-R consistently retrieves turning, backward, and escape maneuvers appropriate for these scenarios, demonstrating strong closed-loop resilience even under severe terrain complexity.

Collectively, these closed-loop evaluations confirm that retrieval-based action generation combined with open-world perception provides robust, adaptive, and collision-resistant behavior, outperforming state-of-the-art action decoders, classifiers, and CNN/DINO backbones across all tested environments.

\begin{figure*}[t]
\centering
\includegraphics[width=0.98\textwidth]{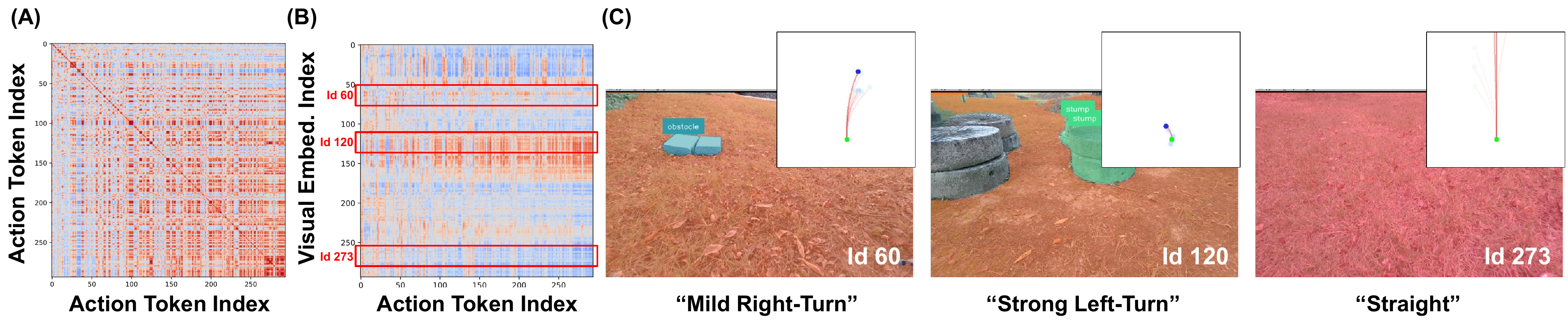}
\vspace{-5pt}
\caption{
(A) Similarity matrix between action tokens. (B) Similarity matrix between streamed visual embeddings and action tokens in real-world experiments. (C) Several example results from the real-world deployment.
}
\label{fig:result_sim}
\vspace{-10pt}
\end{figure*}
\begin{figure*}[t]
\centering
\includegraphics[width=0.98\textwidth]{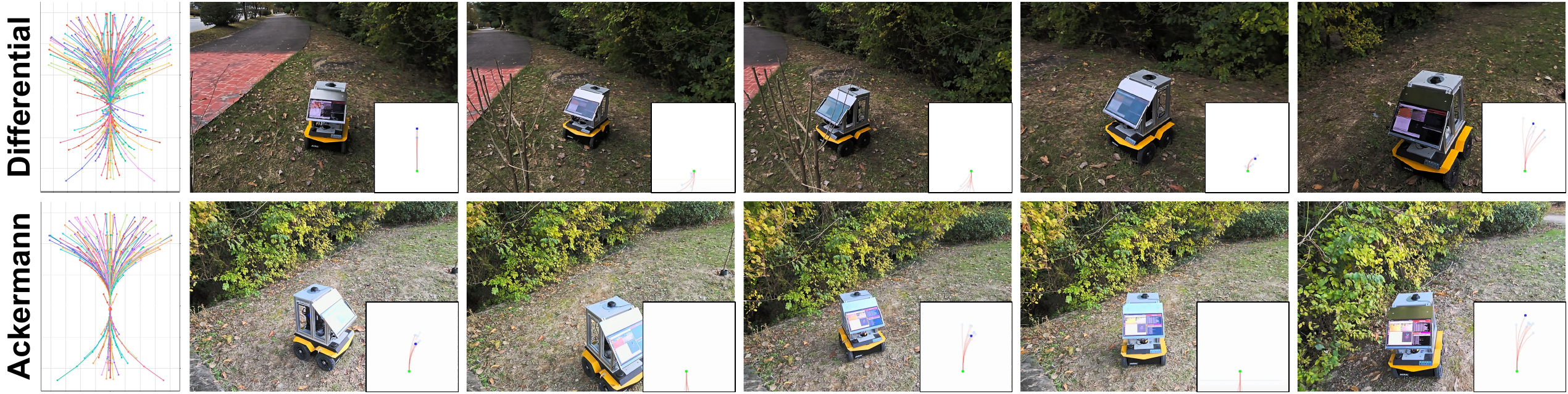}
\vspace{-5pt}
\caption{
Action generalization experiments in the \textit{Dead-End} scenario. Our method generalizes action representation without re-training. We replace the original action token vocabulary for differential-wheeled robots with a different vocabulary designed for an Ackermann steering system, which is incapable of rotational maneuvers.
}
\label{fig:result_action}
\vspace{-10pt}
\end{figure*}

\subsection{Open-World Exploration.}
To evaluate open-world generalization, we conducted real-world exploration in diverse and unfamiliar outdoor environments. Fig. \ref{fig:result_word_cloud} highlights language-level differences between training and deployment, showing that off-road concepts such as ``leaf” and ``stump” appear far more frequently during real-world operation. Fig. \ref{fig:result_vla-r} provides qualitative results across rough-terrain, dense-tree, and cliff-side scenarios, each containing novel visual concepts, irregular geometry, or hard-to-label obstacles unseen during training.
Across all settings, the vision–language backbone provides robust open-world detection, enabling the OW-QFormer to form stable, interpretable scene embeddings under significant appearance shifts. Conditioned on these embeddings, VLA-R retrieves action trajectories that reflect scene semantics: e.g., sharp turns near trees, cautious forward motion in narrow gaps, and back-and-forth maneuvers to escape dead-end or cliff regions. Retrieved actions consistently align with visual cues, showing the model’s ability to adapt in previously unseen situations.
Notably, VLA-R leverages the high-level prompt \textit{“obstacle”} to generalize to novel objects (e.g., stone structures, bricks, stump-like shapes) without explicit supervision, demonstrating its potential for open-world end-to-end autonomous driving.

\subsection{Vision-Language Action Embedding.}
To evaluate whether VLA-R learns a coherent and semantically meaningful joint embedding space, we analyze the alignment between vision–language embeddings produced by the OW-QFormer and the action embeddings encoded by the Action Transformer. Fig. \ref{fig:result_sim} summarizes our findings.

We first examined the structure of the learned action embedding space by computing pairwise cosine similarities between all trajectory tokens. As shown in Fig. \ref{fig:result_sim}(A), the embeddings exhibited clear geometric organization: tokens with similar motion profiles (e.g., mild right turns, strong left turns, straight motions) cluster together, while dissimilar trajectories were well separated. This structure verifies that the Action Transformer produces a smooth and behaviorally interpretable action manifold, rather than a discrete set of isolated motion codes.

We then evaluated cross-modal alignment by streaming real-world perception features into the embedding space. Fig. \ref{fig:result_sim}(B) shows that visual embeddings during real-world deployments consistently activate the correct neighborhood of action tokens, even under significant appearance variations such as dense vegetation, narrow passages, and irregular terrain. The retrieved trajectories (Fig. \ref{fig:result_sim}(C)) closely reflect scene semantics: for example, selecting mild turning motions near brick-like objects, strong rotational maneuvers in cluttered regions, or forward motions when driving collision-free open space covered with \textit{grass} and \textit{leaves}. 
These results demonstrate that VLA-R learns a well-aligned, interpretable vision–action embedding space that supports robust open-world reasoning.

\subsection{Generalizable Action Representation.}
A key advantage of our action retrieval framework is its ability to generalize the underlying action space and representation without re-training. Unlike conventional action decoders or continuous-control policies that bind the learned model to a fixed set of motion primitives or control distributions, VLA-R reasons over a tokenized action representation that can be redefined or reconfigured at inference time. This decoupling enables the policy to adapt to new motion vocabularies and behavioral abstractions, whether coarse, fine-grained, or structurally constrained, while preserving consistent perception–action alignment.

To demonstrate this capability, we evaluated VLA-R by swapping the original action vocabulary, constructed for differential-drive robots, with a newly constructed vocabulary tailored to an Ackermann steering system (Fig. 9). Ackermann platforms exhibit nonholonomic constraints and cannot perform in-place rotational maneuvers, resulting in a fundamentally different set of feasible trajectories. Despite this structural mismatch, VLA-R seamlessly adapts: the policy retrieves appropriate complex, nonholonomic back-and-forth action primitives, without any fine-tuning, indicating robust generalization of the learned vision-action embedding.
Qualitatively, the retrieved actions aligned with vehicle-specific motion constraints: differential-drive tokens produce turning and pivoting behaviors, whereas the Ackermann vocabulary yields smooth arc-like trajectories consistent with front-wheel steering in the narrow dead-end situations. This demonstrates that the shared contrastive embedding captures semantically grounded relationships between perceived scenes and feasible motions, rather than memorizing vehicle-specific control patterns.
These results highlight that VLA-R supports plug-and-play action vocabularies, enabling rapid deployment across heterogeneous robotic platforms and offering a promising pathway toward scalable, generalizable end-to-end autonomy.
\section{Conclusions}

We presented VLA-R, an open-world end-to-end autonomous driving framework that bridges perception and control through a unified vision–language–action embedding space. Leveraging frozen open-world vision-language models and a contrastively aligned action retrieval mechanism, VLA-R achieves generalizable perception, transferable action reasoning, and robust performance in unstructured real-world environments, even with limited data. Our results highlight the potential of vision–language integration to enable scalable, interpretable, and open-world autonomous driving. Although demonstrated on a mobile robot platform, the proposed formulation is general and can extend to urban driving domains with appropriate data, offering a promising direction for future work.

\noindent \textbf{Limitations.}
While our results demonstrate strong performance under limited data, our study was constrained by the use of a small-scale dataset (under two hours of driving). Although VLA-R remains robust, larger and more diverse datasets would further strengthen the alignment between vision–language and action embeddings. In addition, our experiments were conducted in goal-agnostic exploration settings, where the robot rarely revisits the same locations and may exhibit minor fluctuations when approaching centrally located obstacles. Incorporating goal-conditioned information could mitigate these issues and represents a valuable direction for future work.

{
    \small
    \bibliographystyle{ieeenat_fullname}
    \bibliography{main}
}


\end{document}